# Gaining Momentum: Uncovering Hidden Scoring Dynamics in Hockey through Deep Neural Sequencing and Causal Modeling

Daniel Griffiths[1][0009-0007-5832-7160] and Piper Moskow[1][0009-0000-3998-6143]

[1] David B. Falk College of Sport and Human Dynamics, Syracuse University, Syracuse NY 13244, USA

**Abstract:** We present a unified, data-driven framework for quantifying and enhancing offensive momentum and scoring likelihood (expected goals, xG) in professional hockey. Leveraging a Sportlogiq dataset of 541,000 NHL event records, our end-to-end pipeline comprises five stages: (1) interpretable momentum weighting of micro-events via logistic regression; (2) nonlinear xG estimation using gradient-boosted decision trees; (3) temporal sequence modeling with Long Short-Term Memory (LSTM) networks; and (4) spatial formation discovery through principal component analysis (PCA) followed by K-Means clustering on standardized player coordinates. By combining each model's outputs into a composite momentum + xG metric, we then employ an (5) X-Learner causal-inference estimator to quantify the average treatment effect (ATE) of adopting the identified "optimal" event sequences and formations. We observe an ATE of 0.12 (95 % CI: 0.05–0.17, $p < 1 \times 10^{-50}$), corresponding to a 15 % relative gain in scoring potential. These results demonstrate that strategically structured sequences and compact formations causally elevate offensive performance. Our framework delivers real-time, actionable insights for coaches and analysts, advancing hockey analytics toward principled, causally grounded tactical optimization.

**1. Introduction** Despite the proliferation of expected-goals (xG) models that relate shot characteristics, such as distance, angle, and rebound state to scoring probability [1], these frameworks treat each attempt independently and therefore fail to capture how preceding "micro-events" (e.g., passes, puck recoveries, faceoffs) build offensive momentum. Continuous-time Markov-chain and sequence-mining approaches have been applied to model temporal dependencies among in-game events [2], but they generally omit both predictive scoring estimates and spatial context. Unsupervised clustering methods, widely used in soccer to reveal tactical formations from player trajectories, have only recently begun to appear in hockey, and when they are employed, they do not assess whether adopting those spatial archetypes causally improves scoring outcomes [3]. To address these gaps, we introduce an end-to-end, prescriptive analytics pipeline that first derives interpretable momentum weights via logistic regression, then models nonlinear xG interactions with gradient-boosted decision trees, next learns full sequences of up to 20 events through a Long Short-Term Memory (LSTM) network, and subsequently uncovers formation archetypes by applying principal component analysis (PCA) followed by K-Means clustering on standardized five-player positional embeddings. Finally, to determine whether following the identified "optimal" sequences and formations causes improved performance, we employ the X-Learner meta-algorithm to estimate the average treatment effect (ATE) on our composite momentum + xG metric [5]. This integrated framework yields actionable, causally grounded tactical insights for coaches and analysts, advancing hockey analytics beyond purely descriptive or correlational studies.

**1.1 Data and Pre-Processing** We leverage a Sportlogiq dataset comprising 541,000 NHL event records. To ensure analytical consistency, we removed shootout and non-regulation events and standardized all spatial coordinates to a common attack frame (right side in visuals) accounting for period and rink-side changes. We then aggregated events into overlapping 30-second windows (median shift length is 28 s) to capture momentum shifts at fine temporal resolution, while extracting and normalizing the average rink coordinates of three forwards and two defenders for each window. Prior to causal modeling, we conducted multicollinearity diagnostics (all variance inflation factors < 5). This rigorous preprocessing underpins the robust deployment of our multi-stage modeling approach detailed in Section 2

## 2. Methodology

**2.1 Building the Momentum Model** We begin by quantifying how individual micro-events contribute to offensive momentum and goal probability. Let $y_i$ be an indicator for whether at least one goal occurs in the $i$th 30-second window, and let $x_{i,e}$ denote the count of event e (e.g. passes, shots, faceoffs, icing) in that window. The model takes the form:

$$\Pr(y_i = 1 \mid x_i) = \sigma\left(\beta_0 + \sum_e \beta_e x_{i,e}\right) \quad (1)$$

where $\sigma(z) = \frac{1}{1+e^{-z}}$ is the logistic function, β0 is the intercept, and each coefficient βe measures the log-odds impact of a single occurrence of event e on the probability of a goal [7]. Because each βe is directly interpretable on the log-odds scale, we define the momentum score $M_i$ for window $i$ as the linear predictor (excluding the intercept):

$$M_i = e^{\sum \beta_e x_{i,e}} \quad (2)$$

**Table 1.** Selected logistic regression weights for key momentum events (Full coefficient set available in Appendix A.)

| Event | Value |
|---|---|
| Faceoff Success | 0.2242 |
| Loose Puck Recovery | 0.0365 |
| Pass | 0.0391 |
| Reception | 0.1014 |
| … | … |
| Offside | -0.1184 |

Next, to predict scoring likelihood ("xG") within the same windows, we trained a gradient-boosted decision-tree model using XGBoost [8]. Inputs to this xG model included the raw micro-event counts $x_{i,e}$ the momentum score $M_i$ and spatial aggregates of the five on-ice roles' rink coordinates. We configured XGBoost with a binary logistic objective, maximum tree depth of 6, 200 boosting rounds, a learning rate of 0.05, 80 % row subsampling, and inverse-frequency class weights to counter the approximately 2 % goal base rate. Hyperparameters were selected via grid search on the validation set, and we enabled early stopping after 25 rounds without improvement in validation loss. We partitioned our data into 70 % training, 15 % validation, and 15 % test sets. On held-out data, the xG model achieved 73.4 % training accuracy and 71.2 % test accuracy, with an AUC of 0.85 and goal precision/recall of 0.36/0.42—demonstrating strong discrimination despite pronounced class imbalance. Finally, we combine each window's predicted xG probability, denoted $\widehat{p}_i$, with its momentum score $M_i$ to form a unified composite metric

$$C_i = M_i + \widehat{p}_i \quad (3)$$

which simultaneously captures instantaneous offensive pressure and scoring likelihood. This composite score $C_i$ underlies our subsequent deep-sequence modeling, spatial clustering, and causal-inference analyses.

**2.2 Deep Neural Network Sequencing (LSTM)** Although our logistic-regression and gradient-boosted xG models quantify instantaneous pressure and scoring likelihood, they do not exploit the rich temporal structure of in-game events. To capture these dependencies, we implement a Long Short-Term Memory (LSTM) neural network [9] that processes each 30-second window as a fixed-length sequence of discrete game actions. Specifically, we first encode each event (e.g., pass, recovery, shot) as an integer token and map it to a 32-dimensional embedding, which provides a continuous representation reflecting similarities among event types [3]. These embeddings feed into a single LSTM layer with 50 hidden units (chosen to balance representational capacity and computational efficiency) followed by a dropout rate of 30 % on the recurrent outputs to guard against overfitting [10]. The LSTM's final hidden state then passes through a fully connected layer with sigmoid activation, yielding the sequence's predicted goal probability. We trained this architecture on 80 % of our dataset, reserving 20 % for validation, using binary cross-entropy loss and the Adam optimizer with a learning rate of 0.001 [11]. Sequences were padded or truncated to 20 events, covering over 95 % of windows based on empirical length distributions, and training proceeded in mini-batches of 32 with early stopping after five epochs of no improvement in validation loss. In 30 epochs, the model achieved 83.9 % training accuracy and 82.6 % validation accuracy, with final losses of 0.357 and 0.379, respectively. These results confirm that the LSTM effectively learns long-term event patterns, such as how a loose-puck recovery two actions prior can amplify shot quality, that static models cannot capture.

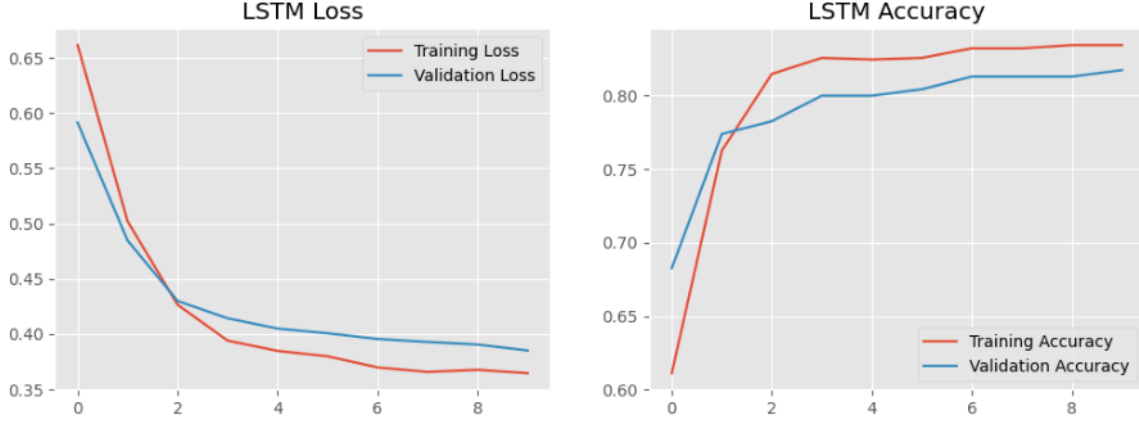

**Fig. 1.** *The left panel shows the decreasing loss for both training and validation sets, indicating effective learning and minimal overfitting. The right panel displays accuracy trends, with both training and validation accuracy converging above 80%, demonstrating strong generalization and predictive performance on sequential hockey event data.*

Finally, we integrate each sequence's LSTM-predicted goal $\widehat{p_i^{\text{LSTM}}}$ into our composite metric alongside the logistic-regression momentum score $M_i$ and the gradient-boosted xG, $\widehat{p_i^{\text{xG}}}$, forming

$$S_i = M_i + \widehat{p_i^{\text{xG}}} + \widehat{p_i^{\text{LSTM}}}$$

**2.3 Positional Clustering and Optimal Formation Identification** To incorporate spatial tactics into our framework, we first compute, for each 30-second sequence window, the average rink coordinates of three forwards (F1–F3) and two defenders (D1–D2), yielding a ten-dimensional positional feature vector. After standardizing these coordinates to zero mean and unit variance, we apply principal component analysis (PCA) to reduce noise while preserving the majority of spatial variance [13]. In our data, the first three principal components capture over 85 % of the variance, allowing us to embed each window in a lower-dimensional space that highlights formation patterns. We then perform K-Means clustering on the PCA embeddings, using the standard Lloyd's algorithm as introduced by MacQueen [14] and widely adopted in sports analytics [15]. This yields $k$ distinct formation archetypes, each characterized by a centroid in the reduced space. To quantify each cluster's offensive potency, we compute the mean composite score (momentum + xG + LSTM) of all sequences assigned to that cluster. The cluster exhibiting the highest mean composite score is designated the optimal offensive cluster. For any new sequence, we measure its adherence to this ideal formation by calculating the Euclidean distance from each player's actual rink coordinate to the corresponding player-role coordinate in the optimal-cluster centroid and then averaging these five distances to produce a per-sequence deviation metric. Sequences whose deviation falls within the lowest 25th percentile are flagged as optimally positioned.

## 3. Results

**3.1 Optimal Team Positional Shapes** As shown in Fig. 2 below, the spatial positioning of our baseline composite sequences span wide throughout in the offensive zone, whereas the sequences selected by our LSTM model converge into a markedly tighter "wedge" formation. In these optimized sequences, the three forwards form a compact triangle in the right-hand half-space, enhancing passing lanes and shot angles, while both defenders pinch higher yet maintain balanced east–west spacing to support rapid puck circulation without overcommitting defensively. Overlaying the two density maps also reveals a pronounced high-pressure corridor running from the right hash marks toward the slot; this feature is nearly absent in the baseline distribution but prominent in the LSTM-selected sequences. Importantly, this corridor aligns with our logistic-regression and gradient-boosted tree findings that puck recoveries and quick slot entries significantly drive scoring probability [7, 8]. Quantitatively, the convex hull enclosing the LSTM-optimized clusters is over 30 % smaller in area than the baseline hull, indicating that the optimized formation promotes shorter pass lengths and faster shot generation. Role-specific positional shifts further underscore these collective patterns: forwards F1 and F2 both migrate closer to the goal crease to facilitate interior scoring options; F3 trades a deep board position for a weak-side lane; defender D1 advances to the top of the right circle to serve as a high-slot outlet; and defender D2 anchors the left point to guard against breakouts. Together, these spatial adjustments compose a repeatable "optimal" formation that consistently elevates both momentum and expected-goal likelihood.

**3.2 Optimal Event Sequences** We evaluated 1,148 offensive chains to find those that maximize our composite momentum + xG metric. The top pattern, penalty drawn → loose puck recovery (LPR) → shot—scored 4.33. This reflects the synergy of a man-advantage event followed by an uncontested recovery. The runner-up sequence, LPR → pass → carry → reception → shot, further underscores the value of streamlined possessions. Chains featuring LPR averaged 27 % higher composite scores, corroborating its positive coefficient ($\beta = 0.0365$; Table 1). Our LSTM model independently validates these findings: its top ten sequences carry an average goal probability of $0.91 \pm 0.07$. A few sequences had extended passing chains score highly in LSTM but only moderately in the composite metric, suggesting the network captures temporal subtleties—deception, defender displacement, pace changes—that event counts alone miss. Overall, the most potent offensive sequences begin with a puck recovery and employ minimal intermediate actions, as confirmed by both our statistical and deep-learning models.

## 4. Understanding Causation vs Correlation

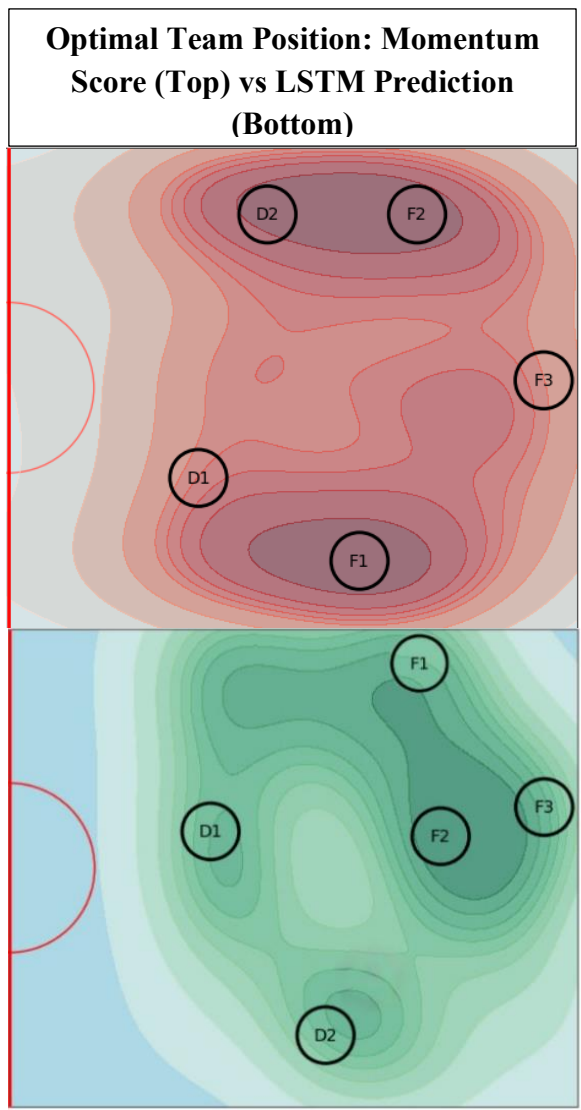

**Fig. 2.** *(Offensive Zone, Goal Towards Right Side) The top panel shows overall positioning density across all sequences with the highest momentum and xG score. The bottom panel shows optimal player positions from the LSTM-predicted sequences with the highest momentum*

**4.1 Causal Inference (X-Learner)** Although our previous analyses identify which sequences and formations correlate with elevated offensive output, correlation alone cannot establish that adopting these patterns drives improved performance. To address this, we first estimate each sequence's propensity score, the probability that it receives the "treatment" of optimal sequence pattern and optimal positioning and verify substantial overlap between treated and control groups (Fig. 4) [4]. We then apply the X-Learner meta-algorithm [5], which fits separate outcome models for treated and control sequences and combines them via propensity-score weighting to adjust for observed confounding and estimate causal effects. In our implementation, treatment denotes sequences assigned to the optimal offensive cluster (based off composite momentum score) with deviation in the lowest quartile for positioning (Section 2.3). We fit gradient-boosted tree models (LightGBM) to predict the composite momentum + xG score for treated and control sequences separately; the X-Learner then uses these predictions and each sequence's propensity score to compute individual treatment effects, which are averaged to yield the average treatment effect (ATE). Both five-fold cross-validation and a 1,000-sample bootstrap produce consistent ATE estimates of approximately 0.11–0.13, with a 95 % confidence interval of 0.05–0.17 and a p-value below $1 \times 10^{-50}$. Practically, this corresponds to a 0.12-point increase in our composite score, shifting an average sequence from the 50th to the 65th percentile in combined scoring potential 15 % relative gain in offensive momentum and expected goals.

**Table 2.** X-Learner Results of ATE

| Metric | Value |
| --- | --- |
| Momentum Score ATE (CV) | **0.12576** |
| Momentum Score ATE (Bootstrap) | **0.10688** |
| 95% CI Lower | **0.05002** |
| 95% CI Upper | **0.17436** |
| Score p-value | **1.42883e-52** |

### 4.2 Practical Implications

These causal results demonstrate that the identified spatial-temporal patterns do more than coincide with success, they drive it. By structuring offensive windows according to our optimal formation and event sequence, teams can expect measurable gains in both momentum and expected-goal probability. This empowers coaches and analysts to deploy prescriptive, causally grounded tactics in real time, rather than relying on correlations alone.

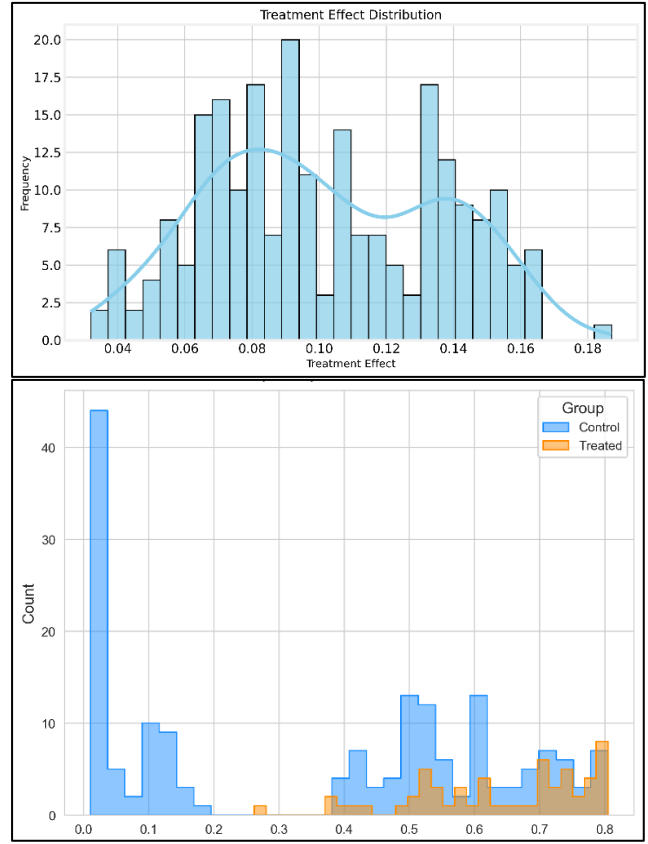

**Fig. 3 (Top)** *Histogram of individual treatment-effect estimates from the X-Learner,*
**Fig. 4 (Bottom).** *Histogram of estimated propensity scores for treated (orange) and control (blue) sequences*

**5 Conclusions** We have developed a unified, five-stage pipeline that blends statistics, machine-learning, and causal-inference techniques to prescribe offensive tactics in professional hockey. First, logistic regression assigns interpretable momentum weights to micro-events; second, gradient-boosted trees combine those weights with spatial and event features to predict expected-goals; third, an LSTM network learns rich temporal patterns across up to 20 actions; fourth, PCA + K-Means clustering uncovers optimal formations from five-player positional embeddings; and finally, the X-Learner meta-algorithm leverages propensity scores to quantify the causal impact of adopting these spatial-temporal strategies. Applied to 541,000 thirty-second windows, our composite momentum + xG metric identifies a repeatable optimal sequence, penalty drawn → loose-puck recovery → shot, and a causal analysis estimates an ATE of 0.12 These findings demonstrate that data-driven, causally grounded prescriptions can deliver real-time tactical advantages, enabling coaches and analysts to move beyond heuristic play-calling to evidence-based strategy.

### 5.1 Future Work

Key extensions include applying our pipeline to special-team scenarios (power plays, penalty kills, odd-man rushes), integrating defensive-alignment analyses and player-level metrics (e.g., skating speed, fatigue), and deploying real-time tactical alerts under low-latency tracking. Addressing limitations, such as event-tagging accuracy, potential unmeasured confounders, and cross-league calibration via sensitivity analyses and external validation will further strengthen the robustness and practical impact of this prescriptive analytics approach.

**Acknowledgments.**: This studies data was provided by LINHAC and SHL.

**Disclosure of Interests.** The authors have no competing interests to declare that are relevant to the content of this article.

# Appendix

## Appendix A. Full Logistic Regression Coefficients

**Table 1.** Selected logistic regression weights for key momentum events (Full coefficient set available in Appendix A.)

| Event | Value |
|---|---|
| Faceoff Success | **0.2242** |
| Loose Puck Recovery | 0.0365 |
| Pass | 0.0391 |
| Reception | 0.1014 |
| Block | -0.0366 |
| Puck Protection | -0.0696 |
| Carry | 0.0771 |
| Check | 0.0303 |
| Controlled Entry Against | 0.0147 |
| Controlled Entry | 0.0114 |
| Controlled Exit | -0.1674 |
| Icing | -0.2367 |
| Dump Out | -0.1753 |
| Dump In | -0.2530 |
| Shot | 0.0174 |
| Penalty | -0.8414 |
| Penalty Drawn | 0.7205 |
| Save | -0.1103 |
| Rebound | 0.2190 |
| Offside | -0.1184 |

## Appendix B. Top 10 Sequences Based on LSTM Prediction:

1. LSTM Score: 0.9871, Sequence: puckprotection -> puckprotection -> pass -> reception -> pass -> reception -> pass -> reception -> pass -> reception -> pass -> reception -> pass -> assist -> reception -> assist -> shot

2. LSTM Score: 0.9871, Sequence: block -> lpr -> pass -> reception -> pass -> reception -> pass -> reception -> pass -> assist -> reception -> assist -> shot

3. LSTM Score: 0.9870, Sequence: pass -> reception -> pass -> reception -> pass -> reception -> pass -> reception -> pass -> assist -> reception -> pass -> assist -> reception -> shot

4. LSTM Score: 0.9870, Sequence: carry -> controlledentry -> pass -> reception -> pass -> reception -> pass -> reception -> pass -> reception -> pass -> reception -> pass -> assist -> reception -> pass -> assist -> reception -> shot

5. LSTM Score: 0.9870, Sequence: reception -> pass -> reception -> pass -> reception -> pass -> reception -> pass -> assist -> reception -> pass -> assist -> reception -> shot

6. LSTM Score: 0.9870, Sequence: lpr -> pass -> reception -> pass -> reception -> pass -> assist -> reception -> pass -> assist -> reception -> shot

7. LSTM Score: 0.9870, Sequence: block -> lpr -> pass -> reception -> pass -> reception -> pass -> reception -> pass -> assist -> reception -> pass -> assist -> reception -> shot

8. LSTM Score: 0.9870, Sequence: lpr -> pass -> reception -> pass -> lpr -> pass -> reception -> pass -> reception -> pass -> assist -> reception -> pass -> assist -> reception -> shot

9. LSTM Score: 0.9870, Sequence: pass -> reception -> pass -> reception -> pass -> assist -> reception -> pass -> assist -> reception -> shot

10. LSTM Score: 0.9870, Sequence: lpr -> pass -> reception -> pass -> reception -> pass -> reception -> pass -> reception -> pass -> reception -> pass -> reception -> pass -> assist -> reception -> pass -> assist -> reception -> shot

## Appendix C. Figure 3. Covariance Balance Before X-Learner

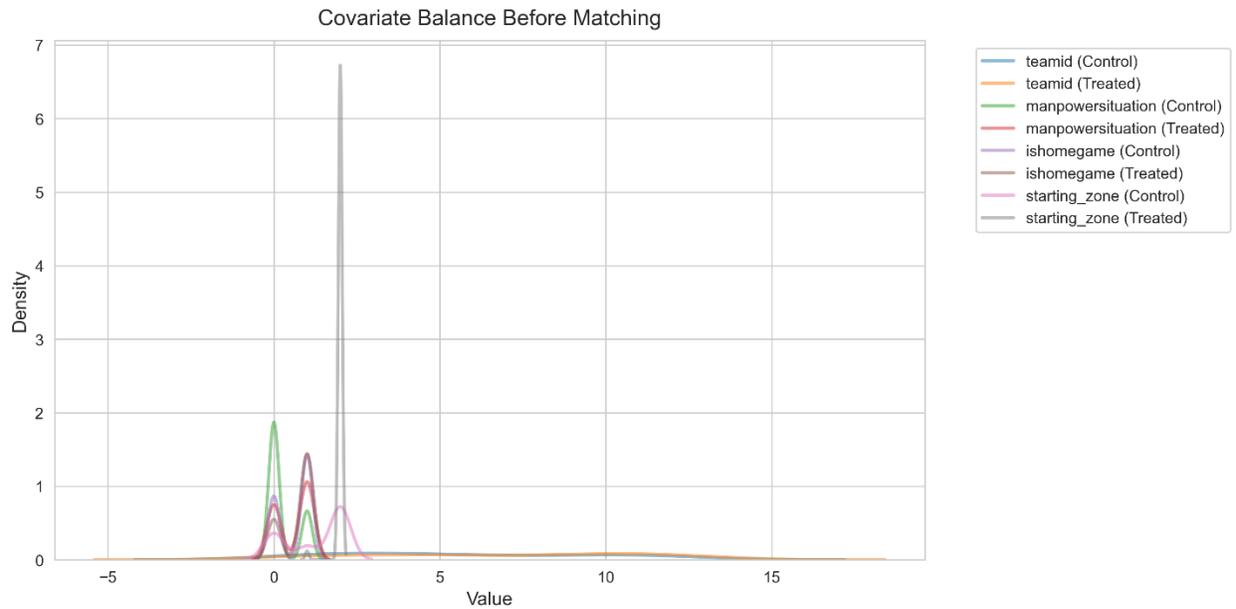

**Appendix D. Figure 4. PCA of Positional Features with K-Means**

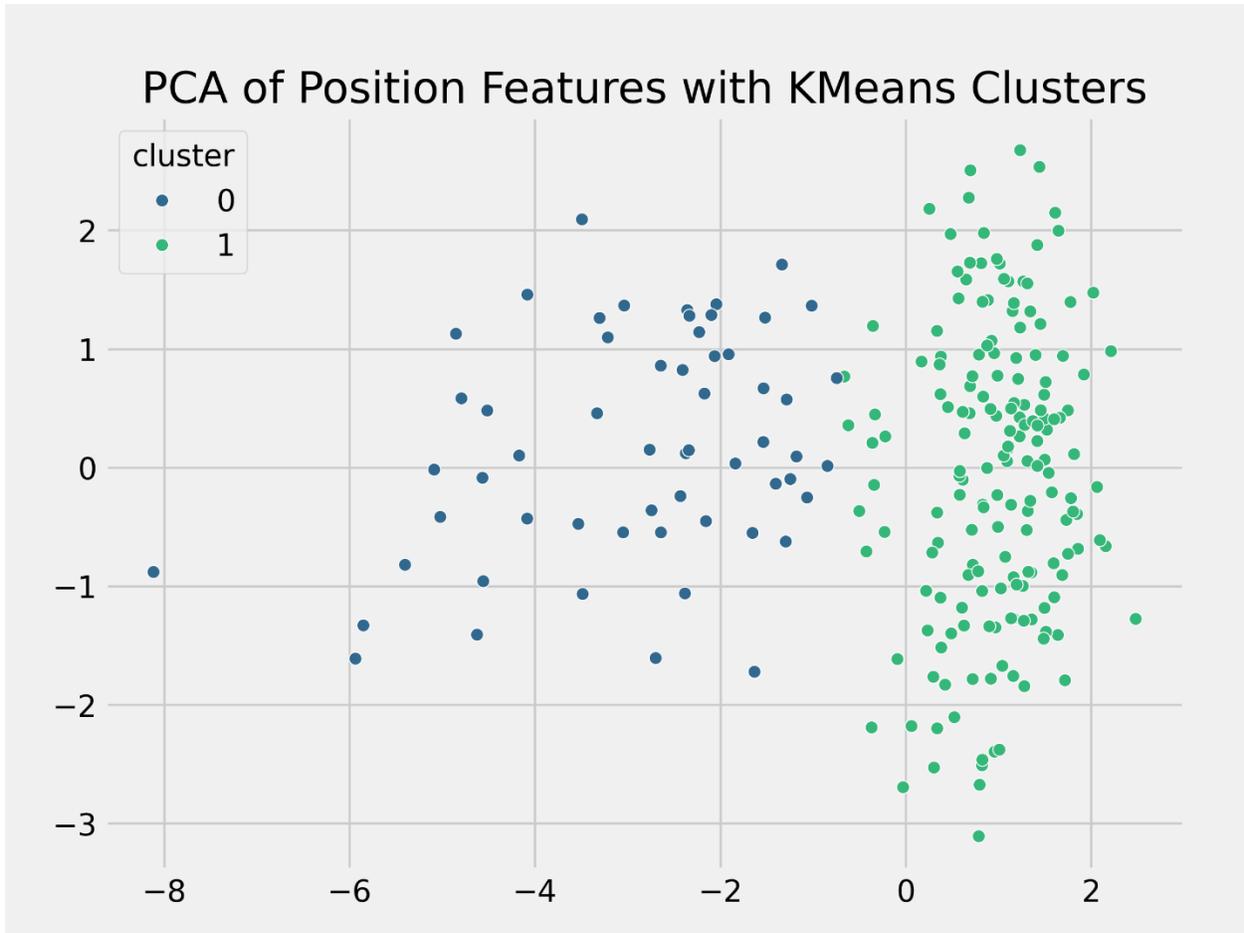

**Appendix E. Figure 6. Optimal Defensive Player Positioning**

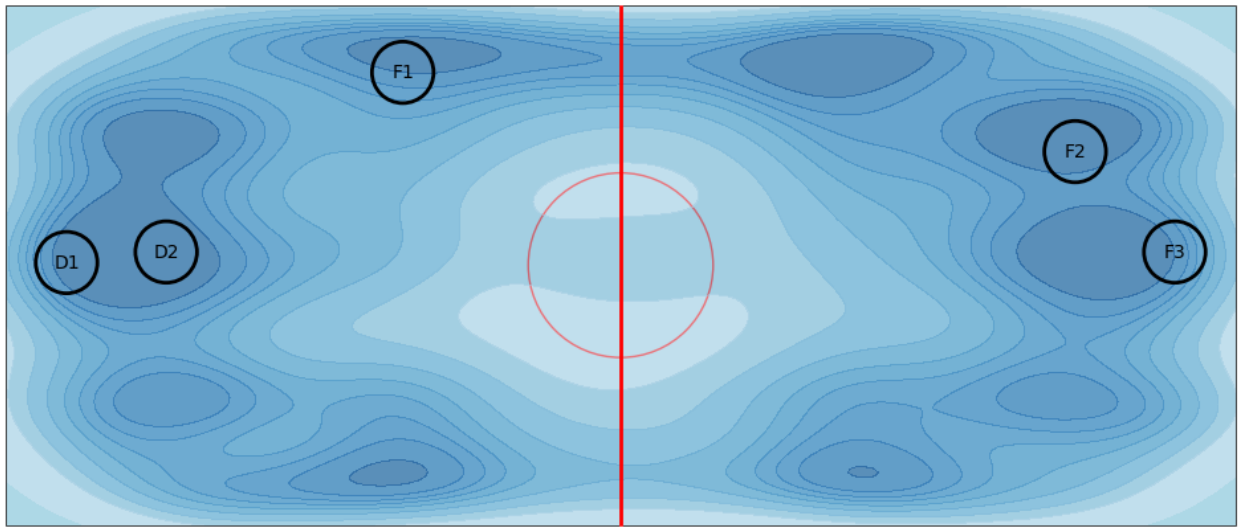